\newcommand{\model}{\textsc{ScreenLLM}\xspace}

\newcommand{\sss}{stateful screen schema\xspace}

\documentclass[sigconf]{acmart}

\usepackage{array}
\usepackage{booktabs}
\usepackage{enumitem}
\usepackage{graphicx}
\usepackage{hyperref}
\usepackage{multirow}
\usepackage{multicol}
\usepackage{amsfonts}
\usepackage{microtype}
\usepackage{xcolor, colortbl}
\usepackage{graphicx}
\usepackage{xspace}

\AtBeginDocument{%
  }

\copyrightyear{2025}
\acmYear{2025}
\setcopyright{cc}
\setcctype{by}
\acmConference[WWW Companion '25]{Companion Proceedings of the ACM Web Conference 2025}{April 28-May 2, 2025}{Sydney, NSW, Australia}
\acmBooktitle{Companion Proceedings of the ACM Web Conference 2025 (WWW Companion '25), April 28-May 2, 2025, Sydney, NSW, Australia}
\acmDOI{10.1145/3701716.3718379}
\acmISBN{979-8-4007-1331-6/2025/04}

\begin{document}

\title{\model: Stateful Screen Schema for Efficient Action Understanding and Prediction}

\author{Yiqiao Jin}
\email{yjin328@gatech.edu}
\authornote{Work done during an internship at Adobe Research.}
\orcid{0000-0002-6974-5970}
\affiliation{%
  \institution{Georgia Institute of Technology}
  \city{Atlanta}
  \state{GA}
  \country{USA}
}

\author{Stefano Petrangeli}
\email{petrange@adobe.com}
\orcid{0000-0002-5492-7747}
\affiliation{%
 \institution{Adobe Research}
 \city{San Jose}
 \state{CA}
 \country{USA}
 }

\author{Yu Shen}
\email{shenyu@adobe.com}
\orcid{0000-0002-9693-0485}
\affiliation{%
  \institution{Adobe Research}
  \city{San Jose}
  \state{CA}
  \country{USA}
 }

\author{Gang Wu}
\email{gawu@adobe.com}
\authornote{Corresponding Author.}
\orcid{0000-0002-8768-2571}
\affiliation{%
  \institution{Adobe Research}
  \city{San Jose}
  \state{CA}
  \country{USA}
}

\renewcommand{\shortauthors}{Yiqiao Jin, Stefano Petrangeli, Yu Shen, and Gang Wu}

\begin{abstract}
Graphical User Interface (GUI) agents are autonomous systems that interpret and generate actions, enabling intelligent user assistance and automation. 
Effective training of these agent presents unique challenges, such as sparsity in supervision signals, scalability for large datasets, and the need for nuanced user understanding. 
We propose \emph{stateful screen schema}, an efficient representation of GUI interactions that captures key user actions and intentions over time. 
Building on this foundation, we introduce \model, a set of multimodal large language models (MLLMs) tailored for advanced UI understanding and action prediction. Extensive experiments on both open-source and proprietary models show that \model accurately models user behavior and predicts actions. Our work lays the foundation for scalable, robust, and intelligent GUI agents that enhance user interaction in diverse software environments. 
\end{abstract}




\maketitle

\section{Introduction}
\label{sec:intro}

GUI agents are autonomous entities that interpret and generate actions in graphical user interfaces (GUIs) to 
automate tasks, understand user behavior, and provide intelligent assistance. 
Recent advancements highlight the potential of these agents.  
From the \emph{data} perspective, the proliferation of multimodal GUI data--such as software usage logs, screen recordings, and voice commands--presents substantial opportunities to enhance these agents' intelligence and contextual awareness. 
YouTube, the second most visited website in the U.S., has over 2.49 billion monthly active users~\cite{YouTubeStats,jin2023predicting}, many of whom rely on tutorial videos on the platform for learning software applications~\cite{rahmatika2021effectiveness, maziriri2020student,li2020screencast}.
From the \emph{model} perspective, multimodal large language models (MLLMs)~\cite{liu2024visual,liu2023improved,luo2024robustft,openai2023gpt4} have shown great promise in processing and understanding multimodal content, making them well-suited for building user agents within GUIs. 


\noindent \textbf{Challenges.} Training GUI agents present several challenges. 
1) \emph{Sparsity in Supervision Signals.} 
Unlike common video data that feature daily activities~\cite{wu2024basketball}, GUIs often remain relatively static, as buttons, menus, and icons show little change unless triggered by user actions. 
This sparsity in supervision signals limits the amount of actionable information that can be extracted from visual content, making agent training difficult. 
\emph{Scalability.} Software usage data such as screen recordings are enormous and require efficient processing methods. Traditional methods, such as frame-by-frame analysis, struggle with scalability and real-time scenarios. 
3) \emph{User Understanding.} To interpret GUI data effectively, agents must comprehend both user actions and underlying intentions, which demands sophisticated human understanding.

\noindent \textbf{This Work.} 
We propose \emph{stateful screen schema}, a compact, informative textual representation of GUI and dynamic user interactions, that captures actions and intentions over time. The schema addresses \emph{data sparsity} and \emph{scalability} by extracting key frames, reducing the need to model all moments in long sessions. 
We then introduce \model, a set of open-source and proprietary models that leverage \emph{stateful screen schema} for UI understanding. 
For open-source MLLMs, we train on screen tutorials--instructional videos that demonstrate software features and workflows. 
Extensive experiments show that \model significantly enhances user understanding and action prediction, such as a 40.7\% improvement in BLEU-2 and 16.5\% in ROUGE-L when using LLaVA-13B as the base model, highlighting its potential for automating complex tasks on GUIs. 


\noindent \textbf{Contributions.} Our contributions are as follows: 
\begin{enumerate}[leftmargin=2em]
    \item \emph{Stateful Screen Schema}: an efficient method to represent GUIs in long user sessions of software usage. 
    \item \emph{ScreenLLM}: A set of Multimodal LLMs tailored for screen UI understanding.
    \item \emph{Empirical Validation}: Experiments on both open-source and proprietary models showing superior performance of \model in understanding user actions and intentions. 
\end{enumerate}

\vspace{-3mm}
\section{Method}
\label{sec:method}

\begin{figure}[t]
    \centering
    \includegraphics[width=0.99\linewidth]{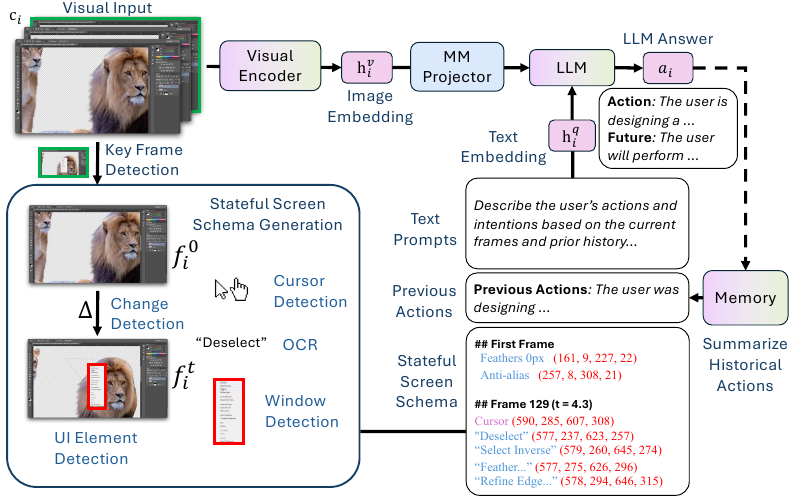}
    \vspace{-3mm}
    \caption{
    \model for understanding user actions and intentions in screen UIs. \model first detects key moments in user actions and generates a \emph{stateful screen schema}, a compact representation of dynamic screen UIs. It then uses the \emph{MLLM} to predict user actions, while the \emph{Memory} module stores and summarizes past actions to support future action understanding and prediction.
    }
    \label{fig:framework}
\vspace{-2mm}
\end{figure}

\subsection{Preliminary}

We model a screen recording video $v$ as a sequence of video clips $\{c_i\}$. Each $c_i$ captures a specific operation $a_i$, such as adding a shape to the design or opening a pop-up window. The goal is to comprehend the user's current action $a_i$ and their intentions (the subsequent action $a_{i+1}$) given a clip $c_i$. 

\vspace{-2mm}
\subsection{Framework Overview}

As illustrated in Figure~\ref{fig:framework}, our \model framework is designed to be highly generalizable and comprises three key components: a \emph{stateful screen schema} generation module, a multimodal LLM, 
and a memory module. 
The stateful screen schema generation module dynamically constructs a structured representation of GUIs and user actions. 
The memory module tracks previous user actions to provide context for inferring current intentions. The multimodal LLM in \model is versatile, supporting both open-source and proprietary MLLMs. 

\begin{table*}[!ht]
    \centering
    \setlength{\tabcolsep}{4pt}
    \begin{tabular}{ll|ccccc|cc}
    \toprule
        Model & Variation & BLEU-1 & BLEU-2 & ROUGE-L & METEOR & CIDEr-D & Category & Tools \\ 
        \midrule
        \multirow{3}{*}{GPT-4o} & ZS & 0.336 & 0.197 & 0.257 & 0.250 & 0.545  & 0.234 & 0.243 \\ 
        & \textbf{SLLM} & \textbf{0.367} & \textbf{0.233} & \textbf{0.276} & \textbf{0.261} & \textbf{0.596} & \textbf{0.319} & \textbf{0.338} \\ 
        & Impr. & 9.2\% & 18.3\% & 7.2\% & 4.4\% & 9.4\% & 36.4\% & 39.1\% \\ 
        \midrule
        \multirow{4}{*}{LLaVA-7B} & ZS & 0.213 & 0.104 & 0.248 & 0.228 & 0.402 & 0.263 & 0.249 \\ 
        & FT & 0.248 & 0.142 & 0.323 & 0.239 & 0.419 & 0.336 & 0.298 \\ 
        & \textbf{SLLM} & \textbf{0.309} & \textbf{0.180} & \textbf{0.348} & \textbf{0.254} & \textbf{0.444} & \textbf{0.416 }& \textbf{0.365} \\ 
        & Impr. & 24.6\% & 26.8\% & 7.7\% & 6.3\% & 6.0\% & 23.6\% & 22.3\% \\ 
        \midrule
        \multirow{4}{*}{LLaVA-13B} & ZS & 0.214 & 0.105 & 0.242 & 0.241 & 0.453 & 0.265 & 0.254 \\ 
        & FT & 0.252 & 0.135 & 0.329 & 0.257 & 0.478 & 0.337 & 0.320 \\ 
        & \textbf{SLLM} & \textbf{0.324} & \textbf{0.190} & \textbf{0.383} & \textbf{0.272} & \textbf{0.545} & \textbf{0.497} & \textbf{0.347} \\ 
        & Impr. & 28.6\% & 40.7\% & 16.5\% & 5.8\% & 14.5\% & 47.4\% & 8.7\% \\ 
        \bottomrule
    \end{tabular}
    \caption{Performance on current action understanding using various base models. \emph{ZS}, \emph{FT}, and \emph{SLLM} stand for the performance in the zero-shot setting, fine-tuned model, and our \textbf{ScreenLLM} framework, respectively. 
    For action description, we use standard image captioning metrics: BLEU-1/2, ROUGE-L, METEOR, and CIDEr-D. For classification tasks like category and tool prediction, we use accuracy.
    }
    \label{fig:action}
    \vspace{-3mm}
\end{table*}



\begin{table*}[!ht]
    \centering
    \setlength{\tabcolsep}{4pt}
    \begin{tabular}{ll|ccccc|cc}
    \toprule
        Model & Variation  & BLEU-1 & BLEU-2 & ROUGE-L & METEOR & CIDEr-D & Category & Tool \\ \midrule
        \multirow{3}{*}{GPT-4o} & ZS & 0.227 & 0.175 & 0.254 & 0.226 & 0.521 & 0.482 & 0.663  \\ 
         & \textbf{SLLM} & \textbf{0.246} & \textbf{0.217} & \textbf{0.283} & \textbf{0.264} & \textbf{0.553} & \textbf{0.492 }& \textbf{0.721}  \\ 
        & Impr. & 8.4\% & 24.0\% & 11.4\% & 16.8\% & 6.1\% & 2.1\% & 8.7\%  \\ 
        \midrule
        \multirow{4}{*}{LLaVA-7B} & ZS & 0.222 & 0.113 & 0.252 & 0.228 & 0.198 & 0.404 & 0.383  \\ 
        & FT & 0.249 & 0.135 & 0.276 & 0.240 & 0.295 & 0.473 & 0.400 \\ 
        & \textbf{SLLM} & \textbf{0.287} & \textbf{0.178} & \textbf{0.327} & \textbf{0.266} & \textbf{0.352} & \textbf{0.498} & \textbf{0.418}  \\ 
        & Impr. & 15.3\% & 31.9\% & 18.5\% & 10.8\% & 19.3\% & 5.3\% & 4.5\%  \\ 
        \midrule
        \multirow{4}{*}{LLaVA-13B} & ZS & 0.224 & 0.114 & 0.316 & 0.233 & 0.184 & 0.425 & 0.395  \\ 
        & FT & 0.266 & 0.150 & 0.332 & 0.257 & 0.463 & 0.451 & 0.420 \\ 
        & \textbf{SLLM} & \textbf{0.324} & \textbf{0.190} & \textbf{0.381} & \textbf{0.271} & \textbf{0.529} & \textbf{0.477} & \textbf{0.442}  \\ 
        & Impr. & 21.8\% & 26.7\% & 14.8\% & 5.4\% & 14.3\% & 5.8\% & 5.2\% \\ 
    \bottomrule
    \end{tabular}
    \caption{Performances on user intention understanding.}
    \label{tab:intention}
    \vspace{-3mm}
\end{table*}


\vspace{-2mm}
\subsection{Stateful Screen Schema Generation}

We propose \emph{stateful screen schema}, a compact textual representation of the screen for each  video segment $c_i$. Unlike previous works focusing on single frames~\cite{baechler2024screenai}, our screen schema models dynamic user sessions.  

\noindent \textbf{Key Frame Extraction.} 
Screen recordings often feature static UIs in the background, making frame-by-frame processing inefficient, especially for real-time AI systems that require prompt responses. 
To improve efficiency, for each clip $c_i$, we extract key frames that capture major UI changes corresponding to important user actions. 
Identifying these key frames is challenging, as different actions lead to different levels of UI changes. For example, zooming into a section of the screen causes significant changes in pixel values across several frames, while actions like typing or adding text to a canvas cause minimal changes.

\begin{figure}[t]
    \centering
    \includegraphics[width=0.99\linewidth]{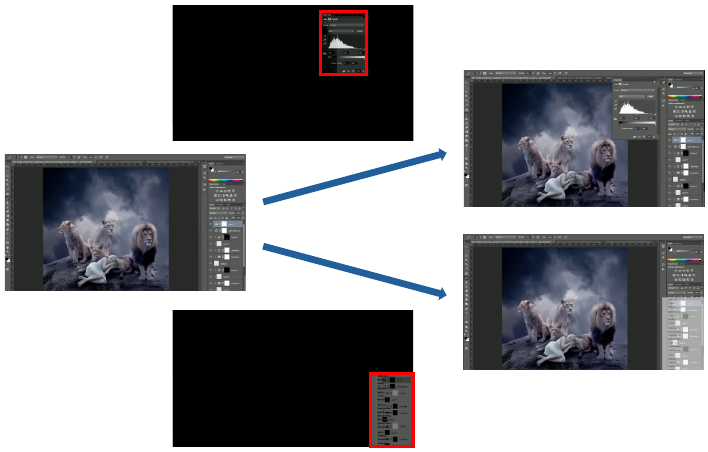}
    \vspace{-3mm}
    \caption{
    The second-order changes in pixel values can be used to detect key frames that involve pop-up windows and menus, corresponding to key user actions. 
    }
    \label{fig:pixel_change}
\vspace{-3mm}
\end{figure}

To capture significant actions, we compute the second-order pixel changes, which measure the variation in pixel differences between adjacent frames. Such changes highlight more pronounced events like the appearance of new windows, menus, or buttons (Figure~\ref{fig:pixel_change}). 
We select the top $k$ frames with the highest second-order pixel changes. ensuring that only frames containing important user interactions are processed, thus improving efficiency. 


\noindent \textbf{UI Element Detection.} 
We identify rectangular areas in the selected frames where pixel changes exceed a threshold $\Delta$ and apply OCR to extract text from these areas. This process helps identify text in various UI elements, such as menus, toolbars, dialog boxes, status messages, as well as content in the workspace such as text layers and user notes.

\noindent \textbf{Cursor Detection.} 
Cursor location reflects the user's focus on the screen. 
We train a compact three-layer CNN model for cursor location detection  
using a diverse dataset that includes screen captures from different operating systems, display resolutions, and application types (e.g., text editors, web browsers, image editors). Data augmentation techniques, including random rotations, scaling, and translations, are employed to improve the model's robustness and simulate various cursor behaviors and real-world user interactions.  

\subsection{Schema Composition}

\begin{figure*}[t]
    \centering
    \includegraphics[width=0.99\linewidth]{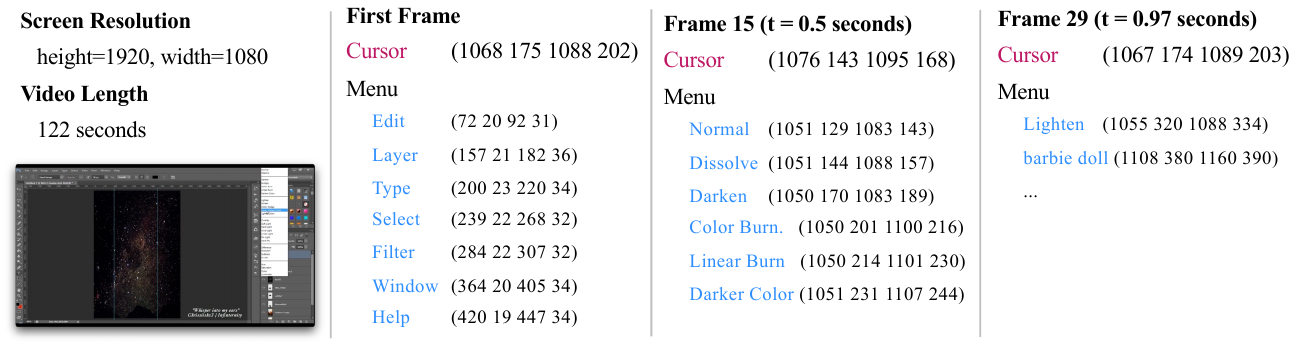}
    \vspace{-3mm}
    \caption{
    A sample stateful screen schema. It consists of the configuration and the initial frame of the video. For the followup frames, we identify the areas with changes, and use OCR to identify them 
    }
    \label{fig:sample_screen_schema}
\end{figure*}

To generate the schema for a clip $c_i$, we first extract all texts and their locations from the first frame $f_i^{0}$ and the identified rectangular regions in subsequent key frames $\{f_i^{t}\}$. The detected texts are then matched with menu items in the software to better understand the user's actions. 
A sample screen schema is shown in Figure~\ref{fig:sample_screen_schema}.
The schema includes information such as timestamps, frame resolution, the locations of cursors and menu items. 
It is then fed into the multimodal LLM alongside a textual prompt and the screenshot of the first frame. This approach tracks changes over time while maintaining a compact and efficient representation of the screen.

\section{Evaluation}
\label{sec:eval}

\noindent \textbf{Metrics.} 
We evaluate our method and the baselines using standard captioning metrics, including METEOR~\cite{banerjee2005meteor},
\ CIDEr-D~\cite{vedantam2015cider}, BLEU~\cite{papineni2002bleu}, and ROUGE-L~\cite{lin2004rouge}.

\noindent \textbf{Dataset.} 
We use the PsTuts dataset~\cite{li2020screencast} for both training and testing. 
PsTuts consists of high-resolution YouTube video tutorials on Adobe Photoshop, with detailed temporal segmentation annotations marking user actions and key transitions between different phases. 
The dataset includes low-level annotations such as selecting tools, opening pop-up windows, drawing shapes, and editing text, from expert Photoshop users sourced from Upwork\footnote{\url{https://www.upwork.com/}}. Annotators provided concise text descriptions and relevant keywords for each temporal segment.
For training, we convert the training samples into instruction-following formats following~\cite{liu2023improved}. 

\noindent \textbf{Implementation Details.} 
We use HuggingFace Transformers~\cite{wolf2020transformers} for model implementation and the PaddleOCR\footnote{\url{https://github.com/PaddlePaddle/PaddleOCR}} Python package for OCR. We set the threshold for bounding box detection to 30. For consistent evaluation results, we use a temperature value of $0.0$ during inference. Where applicable, we set top-$p$ to $0.7$ and the maximum length of the generated text to $256$ tokens.

\noindent \textbf{Task Selection.} Our evaluation includes both classification and natural language generation tasks and focuses on \emph{user action understanding} and \emph{future action prediction}. 
For both tasks, the model must describe he user action in natural language and identify the tools being used. 
These tasks require the model to interpret both low-level visual cues and high-level semantics across multiple frames to understand user actions and intentions.

\begin{table}[ht]
\centering
\setlength{\tabcolsep}{3pt}
\begin{tabular}{@{\hskip 0.02in}ll|cc|cc@{\hskip 0.02in}}
\toprule
& & \multicolumn{2}{c|}{Current Action} & \multicolumn{2}{c}{Future Action} \\
\multicolumn{1}{l}{Model} & Var. & Category & Tool & Category & Tool \\ 
\midrule
\multirow{3}{*}{LLaVA-7B} & ZS & 0.1\% & 4.9\% & 0.3\% & 0.0\% \\ 
& FT & 0.0\% & 1.7\% & 0.0\% & 0.0\% \\ 
& \textbf{SLLM} & \textbf{0.0\%} & \textbf{0.7\%} & \textbf{0.0\%} & \textbf{0.0\%} \\ 
\midrule
\multirow{3}{*}{LLaVA-13B} & ZS& 14.1\% & 26.9\% & 11.2\% & 16.4\% \\ 
& FT & 0.2\% & 8.8\% & 5.3\% & 2.2\% \\ 
& \textbf{SLLM} & \textbf{0.0\%} & \textbf{0.5\%} & \textbf{0.0\%} & \textbf{0.0\%} \\ 
\midrule
\multirow{2}{*}{GPT-4o} & ZS & 23.2\% & 36.1\% & 30.8\% & 23.4\% \\ 
& \textbf{SLLM} & \textbf{8.8\%} & \textbf{15.3\%} & \textbf{15.9\%} & \textbf{17.8\%} \\ 
\bottomrule
\end{tabular}
\caption{Failure rates (percentages of answers that do not satisfy the formatting requirements) in the tasks of current \& future action understanding. Lower values indicate better performance.}
\label{tab:failure_rate}
\end{table}

\begin{table*}[htbp]
\centering
\begin{tabular}{c|p{4.7in}}
\toprule
\textbf{Category} & \textbf{Tools} \\
\midrule
\textbf{Move} & Move Tool, Artboard Tool\\ 
\multirow{2}{*}{\textbf{Marquee}}& Rectangular Marquee Tool, Elliptical Marquee Tool, Single Row Marquee Tool, Single Column Marquee Tool\\ 
\textbf{Lasso}& Lasso Tool, Polygonal Lasso Tool, Magnetic Lasso Tool\\ 
\textbf{Object Selection} & Object Selection Tool, Quick Selection Tool, Magic Wand Tool\\ 
\textbf{Cropping} & Crop Tool, Perspective Crop Tool, Slice Tool, Slice Select Tool\\ 
\textbf{Framing}& Frame Tool \\ 
\textbf{Eyedrop}& Eyedropper Tool, Color Sampler Tool, Ruler Tool, Note Tool, Count Tool \\ 
\multirow{2}{*}{\textbf{Repair}} & Spot Healing Brush Tool, Remove Tool, Healing Brush Tool, Patch Tool, Content-Aware Move Tool, Red Eye Tool\\ 
\textbf{Pen}& Brush Tool, Pencil Tool, Color Replacement Tool, Mixer Brush Tool\\ 
\textbf{Stamp}& Clone Stamp Tool, Pattern Stamp Tool \\ 
\textbf{History Brush}& History Brush Tool, Art History Brush Tool\\ 
\textbf{Eraser} & Eraser Tool, Background Eraser Tool, Magic Eraser Tool\\ 
\textbf{Paint}& Gradient Tool, Paint Bucket Tool \\ 
\textbf{Blur} & Blur Tool, Sharpen Tool, Smudge Tool \\ 
\multirow{2}{*}{\textbf{Anchor}} & Pen Tool, Freeform Pen Tool, Curvature Pen Tool, Add Anchor Point Tool, Delete Anchor Point Tool, Convert Point Tool \\ 
\multirow{1}{*}{\textbf{Type}} & Horizontal Type Tool, Vertical Type Tool, Vertical Type Mask Tool, Horizontal Type Mask Tool\\ 
\multirow{1}{*}{\textbf{Shapes}} & Rectangle Tool, Ellipse Tool, Triangle Tool, Polygon Tool, Line Tool, Custom Shape Tool\\ 
\textbf{Selection}& Path Selection Tool, Direct Selection Tool \\ 
\textbf{Drag} & Hand Tool, Rotate Wheel Tool \\ 
\bottomrule
\end{tabular}
\caption{Photoshop tools are organized into categories based on similar functionalities, with each category encompassing multiple tools that serve related purposes.}
\label{tab:photoshop_tools}
\end{table*}

\vspace{-3mm}
\subsection{Major Findings} 

\noindent \textbf{Overall Performance Improvement.} 
\model (SLLM) consistently and substantially outperforms both zero-shot (ZS) and fine-tuned (FT) settings across tasks (user intention and current action understanding) and metrics (BLEU, ROUGE, METEOR, and CIDEr). 
For \emph{current action understanding} (Table~\ref{fig:action}), the most significant gains are observed on open-source models, particularly LLaVA-13B, where \model reaches a 40.7\% improvement on BLEU-2 and a 16.5\% improvement on ROUGE-L. These results demonstrate \model's ability to effectively adapt to specific tasks when parameter updates are allowed.

\noindent \textbf{Performance on Proprietary Models.} As shown in Table~\ref{tab:intention} \model consistently outperforms the zero-shot baseline on the task of \emph{user intention understanding} with GPT-4o. 
\model improves BLEU-2 by 24.0\% (from 0.175 to 0.217) and CIDEr-D by 6.1\% (from 0.521 to 0.553) compared to the zero-shot setting on GPT-4o. This indicates that 
\sss enhances both the quality of generated text and the classification accuracy even with frozen parameters. 

While the performance improvements on generative tasks are more modest--such as a 2.1\% improvement in category prediction accuracy and an 8.7\% improvement in tool prediction accuracy--the gains in these tasks still suggest that \model significantly enhances the understanding of user actions on screen UIs. 
These findings suggest that the \emph{stateful screen schema} significantly boosts the understanding of large proprietary models on screen UIs, as it provides a structured approach to interpreting the complex visual and textual cues in UI interactions.

\noindent \textbf{Success Rate.} 
The ability of models to follow instructions is important because it directly influences their capacity to generate outputs that align with user expectations and task requirements. 
In tasks such as action understanding or user interface interactions in applications like Photoshop, models must correctly interpret and respond to specific commands or instructions. 
Following previous works~\cite{jin2024mm,lazuka2024llm}, we report the failure ratio (Table~\ref{tab:failure_rate}), defined as the percentage of LLM outputs that fail to meet the requested formats or task specifications. 
Our findings show that applying \model significantly improves the model's ability to produce responses that align with the requested output formats. This improvement suggests that \model enhances the model's instruction-following ability, allowing it to focus more effectively on the task and generate outputs that meet the specified requirements.

\section{Related Works}

\subsection{Large Language Models}
Large language models (LLMs) such as GPT-4 \cite{openai2023gpt4}, LLaMA~\cite{touvron2023llama}, Mistral Large 2~\cite{jiang2023mistral}, and Gemini~\cite{team2023gemini} have shown remarkable generative capabilities~\cite{xiao2023large,xiao2024tradingagents,jin2024agentreview,wang2023parameter}, supporting applications such as virtual assistants and UI agents~\cite{zhang2024ufo,li2024appagent,oh2024uniguard,liu2024tackling,liu2024picture,liu2024timemmd,liu2025can}. 
However, they usually struggle with fine-grained, high-resolution visual data as they often compress input images into low resolutions~\cite{dan2024evaluation,dan2024image,luo2024neural}, which limits their effectiveness for UI elements like small buttons, cursors, and menu items. 
Meanwhile, most LLMs are trained on general-purpose image~\cite{liu2025culturevlm, morshed2023human} and video data~\cite{feng2023video2action,kuehne2011hmdb,dong2024musechat}, which differ significantly from screen recording. This mismatch makes it difficult to apply these models directly to UI-centric videos~\cite{li2020screencast}. 

\subsection{GUI Understanding}
Graphical user interface (GUI) understanding involves parsing user interface screenshots into structured elements to interpret screen semantics and user intentions. This field presents both significant opportunities and challenges in tasks like visual question answering, visual grounding, and screen summarization. 
For example, visual grounding~\cite{deng2021transvg, deng2018visual} aims to map textual plans generated by MLLMs to pixel-level locations on GUIs. 
Screen summarization leverages multimodal models for extracting screen semantics and generating descriptive language, as demonstrated by works like Screen2Words~\cite{wang2021screen2words}, Pix2Struct~\cite{lee2023pix2struct},  OmniParser~\cite{wan2024omniparser}, and Vaquita~\cite{wang2023vaquita}.
In this work, we propose streamlined methods and models for representing screen states to facilitate user intention understanding and action generation. 
\section{Conclusion \& Future Work}
\label{sec:conclusion}

In this work, we propose \emph{stateful screen schema}, a concise and generalizable approach for representing screen UIs. Using this schema, we propose \model, which demonstrates strong performance in action understanding and prediction. 
Future works can expand on this framework to develop GUI agents that support real-time dynamic action generation based on contextual understanding of user actions and intentions.
\section{Ethical Considerations} 

Although we utilize existing licensed dataset such as online videos and annotations for training our GUI framework, training GUI agents generally raises important ethical considerations related to data licensing and user privacy. 

\noindent \textbf{Data Licensing.} Training should only use publicly available and permissible data, adhering to platform policies and intellectual property laws. When applicable, content creators should be credited for their contributions.

\noindent \textbf{User Privacy.}
Identifiable user data should be anonymized or removed during preprocessing. Sensitive or private information, such as user comments or restricted content, should be excluded to ensure compliance with privacy standards and ethical guidelines.

\bibliographystyle{unsrt}
\bibliography{custom}

\end{document}